%% file: en_ieee.tex
\documentclass[10pt,conference]{IEEEtran}
\usepackage{header}
\usepackage{en/preamble}
\usepackage{import}
\usepackage{subcaption}

\author{
    \IEEEauthorblockN{Cleison Correia de Amorim}
    \IEEEauthorblockA{
        \textit{Centro de Informática} \\
        \textit{Universidade Federal de Pernambuco} \\
        50.740-560, Recife, PE, Brazil \\
        cca5@cin.ufpe.br
    }
    \and
    \IEEEauthorblockN{David Macêdo}
    \IEEEauthorblockA{
        \textit{Centro de Informática} \\
        \textit{Universidade Federal de Pernambuco} \\
        50.740-560, Recife, PE, Brazil \\
        dlm@cin.ufpe.br
    }
    \and
    \IEEEauthorblockN{Cleber Zanchettin}
    \IEEEauthorblockA{
        \textit{Centro de Informática} \\
        \textit{Universidade Federal de Pernambuco} \\
        50.740-560, Recife, PE, Brazil \\
        cz@cin.ufpe.br
    }
}

\begin{document}
    \maketitle
    
    \begin{abstract}
    \import{en/}{abstract.tex}
    \end{abstract}
    
    \begin{IEEEkeywords}
    \import{en/}{keywords.tex}
    \end{IEEEkeywords}
    
    \import{en/icann/}{sections.tex}
    
    \bibliographystyle{IEEEtran}
    \bibliography{IEEEabrv,references.bib}
\end{document}

%% file: en/abstract.tex
The recognition of sign language is a challenging task with an important role in society to facilitate the communication of deaf persons. We propose a new approach of Spatial-Temporal Graph Convolutional Network for sign language recognition based on the human skeletal movements. The method uses graphs to capture the dynamics of the signs in two dimensions, spatial and temporal, considering the complex aspects of the language. Additionally, we present a new dataset of human skeletons for sign language based on ASLLVD to contribute to future related studies. 

%% file: en/keywords.tex
Sign Language, Convolutional Neural Network, Spatial Temporal Graph.

%% file: en/icann/sections.tex
\section{Introduction} 
\label{sec:introduction}

Sign language is a visual communication skill that enables individuals with different types of hearing impairment to communicate in society. It is the language used by most deaf people in daily lives and, moreover, the symbol of identification between the members of that community and the main force that unites them~\cite{pereira-choi-2011}.

According to the World Health Organization, the number of people with disabling hearing loss is about 360 million, which is equivalent to a forecast of 1 in 10 individuals around the world~\cite{who-2018}. This data, in turn, highlights the breadth and importance of sign language in the communication of many of these people.

Despite that, there are only a small number of hearing people able to communicate through sign language. This ends up characterizing an invisible barrier that interferes with the communication between deaf and hearing persons, making a more effective integration among these people impossible~\cite{peres-2006}. In this context, it is essential to develop tools that can fill this gap by promoting the integration among the population.

Research related to the sign recognition have been developed since the 1990s, and it is possible to verify significant results~\cite{lim-2016,recent-advances-dl-2017}. The main challenges are primarily related to considering the dynamic aspects of the language, such as movements, articulations between body parts and non-manual expressions, rather than merely recognizing static signs or isolated hand positions.

In this work, we adopt a new approach to performing the recognition of human actions based on spatial-temporal graphs called Spatial-Temporal Graph Convolutional Networks (ST-GCN) \cite {st-gcn-2018}. Using graph representations of the human skeleton, it focuses on body movement and the interactions between its parts, disregarding the interference of the environment around them. Besides, it addresses the movements under the spatial and temporal dimensions, and this allows us to capture the dynamic aspects of the actions over time. These characteristics are consistent to dealing with the challenges and peculiarities of sign language recognition.

The main contributions of the paper are: 1) the proposition of a new technique to sign recognition based on human movement, which considers different aspects of its dynamics and contributes to overcoming some of the main field challenges; and 2) the creation of a new dataset of human skeletons for sign language, currently non-existent, which aims to support the development of studies in this area.

Section~\ref{sec:related-work}~and~\ref{sec:st-gcn} present the related work and the details of ST-GCN. In Section~\ref{sec:new-dataset},  the creation of the new database is discussed.
Section~\ref{sec:st-gcn-for-sl-recognition} addresses the adjustments in ST-GCN and Section~\ref{sec:experiments} the conduction of the experiments. Finally, Sections~\ref{sec:results}~and~\ref{sec:final-remarks} contain the results and final remarks, respectively.

\section{Related work}
\label{sec:related-work}

The recognition of sign languages obtained significant progress in recent years, which is motivated mainly by the advent of advanced sensors, new machine learning techniques, and more powerful hardware~\cite{recent-advances-dl-2017}. Besides, approaches considered intrusive and requiring the use of sensors such as gloves, accelerometers, and markers coupled to the body of the interlocutor have been gradually abandoned and replaced by new approaches using conventional cameras and computer vision techniques.

Due to this movement, it is also notable the increase in the adoption of techniques for feature extraction such as SIFT, HOG, HOF and STIP to preprocess images obtained from cameras and provide more information to machine learning algorithms~\cite{laptev-2008,lim-2016}.

Convolutional Neural Networks (CNN), as in many computer vision applications, obtained remarkable results in this field with accuracy reached 90\% depending on the dataset \cite{taskiran-2018,rao-2018}. There are still some variations as 3D CNNs, the combination with other models such as Inception or the Regions of Interest applications~\cite{elbadawy-2017,das-2018,sajanraj-2018}. Recurrent Neural Networks and Temporal Residual Networks also obtained interesting results in the same purpose~\cite{konstantinidis-2018,pigou-2017}.

Despite the above advances, a large portion of these studies addresses static signs or single-letter images, from the dactylology\footnote{Dactylology - digital or manual alphabet, generally used by the deaf to introduce a word that does not yet have an equivalent sign~\cite{pereira-choi-2011}.
}~\cite{taskiran-2018,das-2018}. The problem is the negative effect on the intrinsic dynamics of the language, such as its movements, non-manual expressions, and articulations between parts of the body~\cite{quadros-2004}. In this sense, it is extremely relevant that new studies observe such important characteristics.

With this purpose, we present an approach based on skeletal body movement to perform sign recognition. This technique is known as Spatial-Temporal Graph Convolutional Network (ST-GCN)\footnote {Available at \url{https://github.com/yysijie/st-gcn}.} and was introduced in~\cite{st-gcn-2018}. The approach aims for methods capable of autonomously capturing the patterns contained in the spatial configuration of the body joints as well as their temporal dynamics.

\section{Spatio-Temporal Graph Convolutional Networks}
\label{sec:st-gcn}

\begin{figure}
    \centering
    \begin{subfigure}{.30\textwidth}
        \centering
        \includegraphics[height=2.7cm]{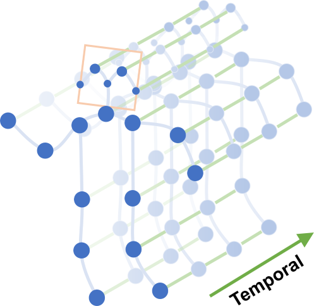}
        \caption{}
        \label{fig:st-gcn-graph}
    \end{subfigure}
    \begin{subfigure}{.30\textwidth}
      \centering
      \includegraphics[height=2.7cm]{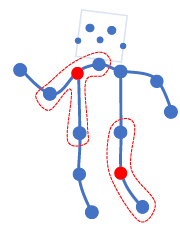}
      \caption{}
      \label{fig:st-gcn-sampling}
    \end{subfigure}
    \begin{subfigure}{.30\textwidth}
      \centering
      \includegraphics[height=2.7cm]{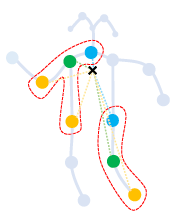}
      \caption{}
      \label{fig:st-gcn-spatial-part}
    \end{subfigure}
    \caption{
        (\subref{fig:st-gcn-graph})~Sequence of skeleton graphs, denoting human movement in space and time. 
        (\subref{fig:st-gcn-sampling})~Sampling strategy in a convolution layer for a single frame.
        (\subref{fig:st-gcn-spatial-part})~Spatial Configuration Partitioning strategy \cite{st-gcn-2018}.
    }
    \label{fig:graph_part_sampling}
\end{figure}

The ST-GCN uses as a base of its formulation a sequence of skeleton graphs representing the human body obtained from a series of action frames of the individuals. Figure~\ref{fig:st-gcn-graph} shows this structure, where each node corresponds to an articulation point. The intra-body vertices are defined based on the body's natural connections. The inter-frame vertices, in turn, connect the same joints between consecutive frames to denote their trajectory over time~\cite{st-gcn-2018}. 

\begin{figure*}[!ht]
    \centering
    \includegraphics[width=1.0\textwidth]{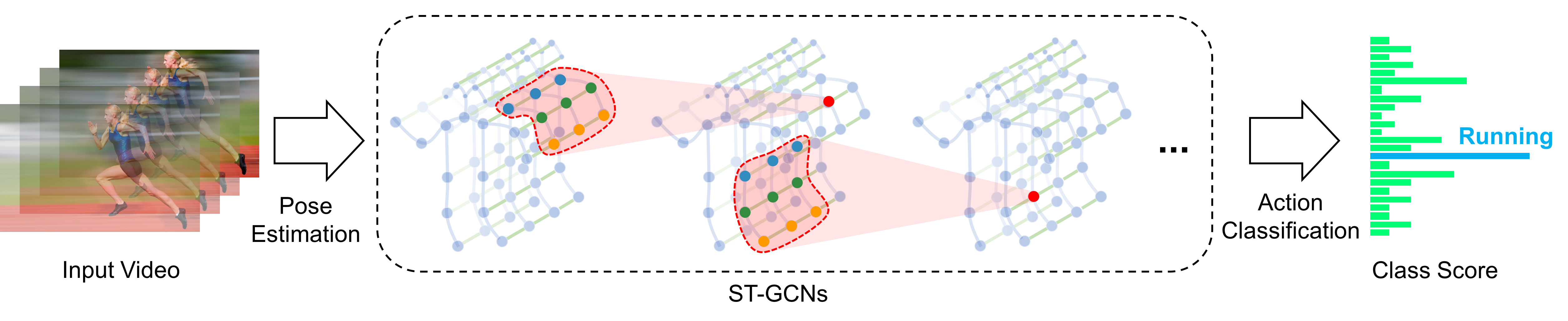}
    \caption{Overview of the ST-GCN approach \cite[p. 3]{st-gcn-2018}.}
    \label{fig:st-gcn-workflow}
\end{figure*}

Figure \ref{fig:st-gcn-workflow} gives an overview of this technique. First, the estimation of individuals' skeletons in the input videos, as well as the construction of space-time graphs based on them. Then, multiple ST-GCN convolution layers are applied, gradually generating higher and higher levels of feature maps for the presented graphs. Finally, they are submitted to a classifier to identify the corresponding action.

To understand the operation of the ST-GCN, it is necessary first to introduce its sampling and partitioning strategies. When we are dealing with convolutions over 2D images, it is easy to imagine the existence of a rigid grid (or rectangle) around a central point that represents the sampling area of the convolutional filter, which delimits the neighborhood. In the case of graphs, however, it is necessary to look beyond this definition and consider the neighborhood of the center point as points that are directly connected by a vertex. Figure \ref{fig:st-gcn-sampling} represents this definition for a single frame. Note that for the red center points, the dashed edges represent the sampling area of the convolutional filter. Note also that although there are other points physically close to the central points (such as the points of the feet, knees, and waist), the method does not consider these points unless there are vertices connecting them to the red ones. This sequence of steps is the \textbf{sampling strategy} of ST-GCN. 

Figure~\ref{fig:st-gcn-sampling} shows how the convolutional filter considers only points immediately connected to the central points. In other words, the delimitation of the filter area considers the neighbors with distance $D = 1$. The authors defined this distance in ST-GCN proposition paper \cite{st-gcn-2018}. 

The \textbf{partitioning strategy} is based on the location of the joints and the characteristics of the movement of the human body, as shown in Figure~\ref{fig:st-gcn-spatial-part}. According to the authors, concentric or eccentric movements categorize the body parts, and the points in the sampling region are partitioned into three subsets: 
    
\begin{itemize}
    \item The root node (or center point, marked green in Figure~\ref{fig:st-gcn-spatial-part});
    \item The centripetal group (blue dots in the Figure~\ref{fig:st-gcn-spatial-part}), which are the neighborhood nodes that are closest to the center of gravity of the skeleton (black cross);
    \item The centrifugal group (yellow dots in the Figure~\ref{fig:st-gcn-spatial-part}), which are the nodes farther from the center of gravity.
\end{itemize}

The center of gravity is taken to be the average coordinate of all joints of the skeleton in one frame. During convolution, each point of the body is labeled according to one of the above partitions named Spatial Configuration Partitioning~\cite{st-gcn-2018}. It is through this method that the authors also establish the weights of the model, making each partition to receive a different weight to be learned.

To learn the temporal dimension, the ST-GCN extends the concept of graph convolution shown above, considering this dimension as a sequence of skeletons graphs stacked consecutively, as in Figure~\ref{fig:st-gcn-graph}. With this, we have in hands a set of graphs that are neighbors to each other. Let us now assume that each articulation of the body in a graph must be connected using a vertex to itself in the graph of the previous neighbor frame and also in the next neighbor frame. Given that, if we return to the definition of sampling introduced above, we verify that the convolutional filter contemplates those points belonging to the neighboring graphs, that now fit the requirements of being directly connected at a distance $D = 1$. In this way, ST-GCN considers the spatial and temporal dimensions and applies convolutions on them.

The OpenPose library was used by~\cite{st-gcn-2018} to estimate the skeletons of individuals in videos. It is an open-source tool that uses deep learning algorithms to detect and estimate up to 130 human body points, as presented in~\cite{cao-realtime-2017,simon-hand-2017,wei-cpm-2016}.

\section{New dataset of human skeletons for sign language} 
\label{sec:new-dataset}

We introduce a new dataset of human skeletons for sign language based on the American Sign Language Lexicon Video Dataset (ASLLVD). The ASLLVD  is a broad public dataset\footnote{ Available at \url{http://csr.bu.edu/asl/asllvd/annotate/index.html}.} containing video sequences of thousands of American Sign Language (ASL) signs, as well as their annotations, respective start point, and end frame markings, and class labels for each sample~\cite{athitsos-asllvd-2008,neidle-2012}.

\begin{figure*}[!ht]
    \centering
    \includegraphics[width=0.9\textwidth]{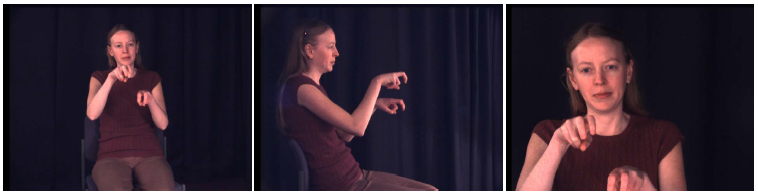}
    \caption{Example of the sign "MERRY-GO-ROUND" from three different perspectives, in the ASLLVD \cite[p. 2]{athitsos-asllvd-2008}.}
    \label{fig:asllvd-example}
\end{figure*}

According to the authors, each sign in ASLLVD is articulated by native individuals in ASL, and video sequences are collected using a four-camera system that simultaneously captures two frontal views, one side view and one enlarged view of the face of these individuals. Figure~\ref{fig:asllvd-example} exemplifies capturing three of these views for the "MERRY-GO-ROUND" sign. 

There is a total of 2,745 signs, represented in approximately 10 thousand samples. Each sign contains between 1 and 18 samples articulated by different individuals (the average number of samples per sign is 4). 

\begin{figure*}[ht!]
    \centering
    \includegraphics[width=0.9\textwidth]{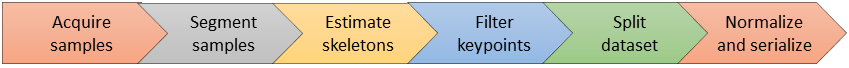}
    \caption{Preprocessing steps for creating the ASLLVD-Skeleton dataset.}
    \label{fig:preprocessamento}
\end{figure*}

Figure~\ref{fig:preprocessamento} shows a series of preprocessing steps to make the ASLLVD samples compatible with the ST-GCN model input.  These steps, in turn, gave rise to a new dataset consisting of the skeletal estimates for all the signs contained therein, which was named \textbf{ASLLVD-Skeleton}\footnote{
   Available at \url{http://www.cin.ufpe.br/~cca5/asllvd-skeleton}
}.

The first step consists of \textbf{obtaining the videos} based on the associated metadata file used to guide this process. We consider only the videos captured by the frontal camera, once they simultaneously contemplate movements of the trunk, hands, and face of the individuals.

The next step is to \textbf{segment the videos} to generate a video sample for each sign. Once every file in the original dataset corresponds to multiple signs per individual, it is necessary to change this organization in such a way that each sign is arranged individually with its label. The output of this step consists of small videos with a few seconds.

\begin{figure}
    \centering
    \begin{subfigure}{3.5cm}
        \centering
        \includegraphics[width=3.5cm]{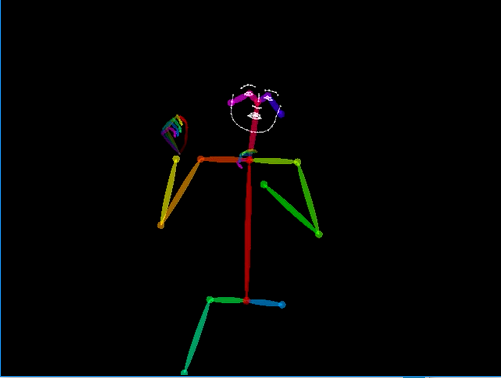}
        \caption{}
        \label{fig:sign-pose-skeleton}
    \end{subfigure}
    \begin{subfigure}{3.5cm}
      \centering
      \includegraphics[width=3.5cm]{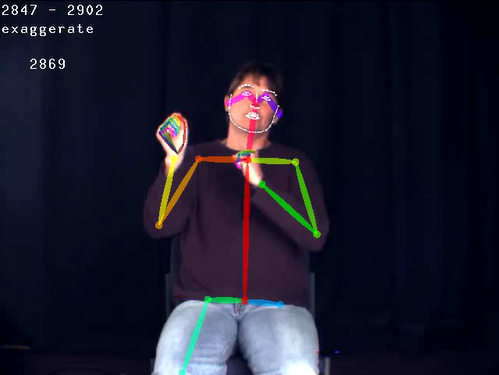}
      \caption{}
      \label{fig:sign-pose-blended}
    \end{subfigure}
    \caption{
        Reconstruction of the skeleton from the coordinates estimated by OpenPose for the sign "EXAGGERATE"~(\subref{fig:sign-pose-skeleton}); and the overlaping this skeleton in the original video~(\subref{fig:sign-pose-blended}).
    }
    \label{fig:sign-pose}
\end{figure}

The third step consists of \textbf{estimating the skeletons} of the individuals in segmented videos. In other words, the coordinates of the individuals' joints are estimated for all frames, composing the skeletons that can be used to generate the graphs of the ST-GCN method. As in \cite {st-gcn-2018}, we used the OpenPose library in this process, and a total of 130 key points were estimated. Figure~\ref{fig:sign-pose} illustrates the reconstruction in a 2D image of the estimated coordinates for the "EXAGGERATE" sign. 

\begin{figure}
    \centering
    \begin{subfigure}{3.0cm}
        \centering
        \includegraphics[height=3.2cm]{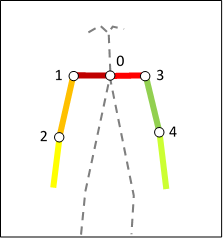}
        \caption{}
        \label{fig:filtered-keypoints-body}
    \end{subfigure}
    \begin{subfigure}{3.0cm}
      \centering
      \includegraphics[height=3.2cm]{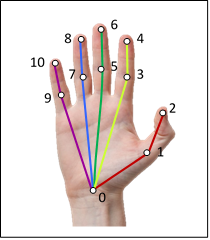}
      \caption{}
      \label{fig:filtered-keypoints-hand}
    \end{subfigure}
    \caption{
        Representation of the 27 used key points, which 5 refer to the shoulders and arms~(\subref{fig:filtered-keypoints-body}) and 11 refer to each hand~(\subref{fig:filtered-keypoints-hand}).
    }
    \label{fig:filtered-keypoints}
\end{figure}

The fourth step involves \textbf{filtering the key points}. We use only 27 of the 130 estimated key points, which 5 refer to the shoulders and arms, and 11 refer to each hand, as illustrated in Figure \ref{fig:filtered-keypoints}. 

The fifth step concerns the \textbf{division of the dataset} into smaller subsets for training and test. For this procedure, a cross-validation dataset tool called "\textit{train\_test\_split}", available by the Scikit-Learn~\cite{scikit-learn} library was used. In this division, we assign a proportion of 80\% of the samples for training (corresponding to 7,798 samples) and 20\% for tests (corresponding to 1,950 samples). This division is a commonly adopted proportion, and we understand that the number of samples in these groups is enough to validate the performance of most machine learning models.

Finally, the sixth step is to \textbf{normalize and serialize} the samples to make them compatible with the ST-GCN input. Normalization aims to make the length of all samples uniform by applying the repetition of their frames sequentially to the complete filling of an established fixed number of frames. The number of fixed frames adopted here is 63 (using a rate of 30 FPS, it is equivalent to a video with a duration of approximately 2 seconds). Serialization, in turn, consists of preloading the normalized samples and translating them into physical Python files which contain their in-memory representation, as in ST-GCN \cite{st-gcn-2018}. We adopted this format to optimize the data loading process. 

The source code that performs these steps is also available in the ASLLVD-Skeleton download area$^5$.

\section{ST-GCN for sign language recognition} 
\label{sec:st-gcn-for-sl-recognition}

Since the graph representation approach adopted by the ST-GCN is very flexible, it is not necessary to make modifications in the architecture of the model. Instead, only specific adaptations to consider the new coordinates of the sign language domain were needed.

The first one is to modify the algorithm to make it able to use new customized layouts of graphs beyond those defined initially in~\cite{st-gcn-2018}. Thus, a new type of layout called "custom" was defined in the configuration file of the model. Besides, a new parameter called "custom\_layout" has been created within the "graph\_args" attribute to allow the number of nodes, the central node, and the edges of new graphs to be informed.

With this configuration, it is possible to map the topology of the graph used in this paper, which is composed of the 27 joints and vertices shown in  Figure~\ref{fig:filtered-keypoints}. Finally, small adjustments were required in the dimensions of the matrices used by the data feeders and in the file
that defines the graph domain of the model so that they can now support layouts with dynamic dimensions.

The source code containing these adaptations is public available\footnote{
    Available at \url{http://www.cin.ufpe.br/~cca5/st-gcn-sl}
}. It consists of a \textit{forked} repository created from the one developed by the authors in~\cite{st-gcn-2018}.

\section{Experiments} 
\label{sec:experiments}

We used as reference the experiments proposed in~\cite{lim-2016}, which evaluate the performance of models as the Block-Based Histogram of Optical Flow (BHOF) (see section \ref{sec:related-work}) and popular techniques such as Motion Energy Image (MEI), Motion History Image (MHI), Principal Component Analysis (PCA), and Histogram of Optical Flow (HOF) in the ASLLVD dataset.

The authors used a subset containing 20 signs selected from the ASLLVD, as presented in Table~\ref{tab:asllvd-20}. To reproduce this configuration, we selected the estimated skeletons for these signs from the ASLLVD-Skeleton dataset. 
\begin{table}[ht]
\centering
\caption{Selected signs for the experiments in~\cite{lim-2016}.}
\label{tab:asllvd-20}
\begin{tabular}{ l | p{0.8\linewidth} }
\hline
Dataset & Selected signs \\ \hline
ASLLVD & adopt, again, all, awkward, baseball, behavior, can, chat, cheap, cheat, church, coat, conflict, court, deposit, depressed, doctor, don’t want, dress, enough \\ \hline
\end{tabular}
\end{table}

We also identified that among the selected signs, there were different articulations for the signs AGAIN, BASEBALL, CAN, CHAT, CHEAP, CHEAT, CONFLICT, DEPRESS, DOCTOR, and DRESS. To solve this, we chose to keep only the articulation that contained the most significant number of samples for their signs.

Since the number of samples resulting from this process was small, totaling 131 items, it was necessary to apply a new division making 77\% of this subset for training and 33\% for tests. With this strategy, we improved the balance in the number of samples for an adequate evaluation of the model. Finally, as we have a new sample division, it was also necessary to normalize and serialize the samples to be compatible with the ST-GCN. 
The resulting subset was named \textbf{ASLLVD-Skeleton-20}\footnote{ Available at \url{http://www.cin.ufpe.br/~cca5/asllvd-skeleton-20} }. 

Due to the characteristics of this small dataset and based on preliminary experiments, the size of the used batch is of 8 samples. In addition, we used a learning rate of 0.01 and the Adam optimizer.

Besides, an experiment with the complete ASLLVD dataset was also conducted to establish a reference value. In this experiment, we set the batch size to 24. 
The learning rate and optimizer were the same as in the experiment above.

\section{Results} 
\label{sec:results}

\begin{figure*}[!t]
    \centering
    \includegraphics[width=0.85\textwidth]{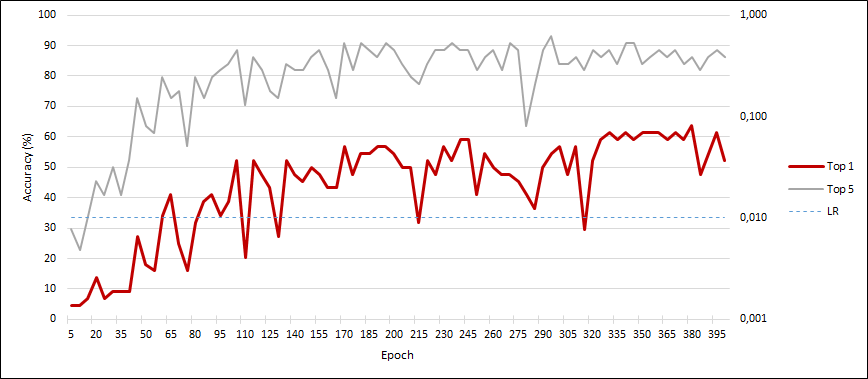}
    \caption{Accuracy obtained by the presented approach in recognition of the 20 signs selected from the ASLLVD.}
    \label{fig:training-asllvd-20}
\end{figure*}

Figure~\ref{fig:training-asllvd-20} presents the first experiment using the approach presented in~\cite{lim-2016}, which considers the selection of 20 specific signs of the ASLLVD. The red line presents the accuracy of the model (\textit{top-1}) and its evolution throughout training epochs. The gray line represents the \textit{top-5} accuracy, which corresponds to the accuracy based on the five most likely responses presented by the model. Finally, the blue dashed line represents the evolution of the learning rate used in the respective epochs and its decay behavior.

From the image, we can observe that the model was able to achieve an average accuracy of 61.04\% between the epochs 330 and 380 in sign recognition. The respective average \textit{top-5} accuracy, in turn, was of 86.36\%. This performance was superior to the results presented by traditional techniques such as MEI, MHI, and PCA, but was not able to overcome that obtained by the HOF and BHOF techniques~\cite{lim-2016}. Table~\ref{tab:results-comparison-20} presents the comparison of these results.

\begin{table}[ht]
\centering
\caption{Sign recognition accuracy using different approaches as proposed in~\cite{lim-2016}.}
\label{tab:results-comparison-20}
\begin{tabular}{lc}
\hline
                   & Accuracy (\%)  \\ \hline
MHI                & 10.00                     \\
MEI                & 25.00                     \\
PCA                & 45.00                     \\
\textbf{ST-GCN SL} & \textbf{61.04}            \\
HOF                & 70.00                     \\
BHOF               & 85.00                     \\ \hline
\end{tabular}
\end{table}

We performed a second experiment using 2,745 signs to establish a reference to the complete ASLLVD dataset. In this scenario, an accuracy (\textit{top-1}) of 16.48\% and a \textit{top-5} accuracy of approximately 37.15\% was obtained with 1,400 epochs. Of course, this is a much more challenging task than the one proposed in~\cite{lim-2016}, and these results reflect this complexity.

From the table, we can see that the approach presented in this paper, based on graphs of the coordinates of human articulations, has not yet been able to provide such remarkable results as that based on the description of the individual movement of the hands through histograms adopted by BHOF. Indeed, the application of consecutive steps for optical flow extraction, color map creation, block segmentation and generation of histograms from them were able to ensure that more enhanced features about the hand movements were extracted favoring its sign recognition performance. This technique is derived from HOF and differs only by the approach of focusing on the hands of individuals while calculating the optical flow histogram.

Methods such as MEI and MHI, however, present more primitive approaches, which mainly detect the movements and their intensity from the difference between the consecutive frames of actions. They are not able to differentiate individuals or to focus on specific parts of their body, causing movements of any nature considered equivalently. The PCA, in turn, adds the ability to reduce the dimensionality of the components based on the identification of those with more significant variance and that, consequently, is more relevant for movement detection in the frames.

The ASLLVD-Skeleton database, in turn, presented high relevance and was able to comply with its purpose of making feasible and supporting the development of this work.  Through its approach and format, it allowed the benefits of adopting a robust and assertive technique to read human skeleton without restrictions or performance penalty during experiments. Considering that such techniques usually have a high computational cost, especially when combined with complex deep learning models, this is a factor that commonly restricts or impedes the progress of researches that seeks to evolve in this direction, based on the coordinates of the body. Thus, this paper contributes and extends to the academic community the benefits found here, encouraging and enabling other advances in sign language recognition.

\section{Final remarks} 
\label{sec:final-remarks}

The results obtained with the presented approach did not reach such expressive performance as those obtained by some of the related techniques. However, its contributions are significant in guiding the next steps to be taken by future studies in this field.

Based on our observations, it is relevant to seek approaches capable of enriching the information about the movements of the estimated coordinates, especially those of the hands and fingers, which, although very subtle, play a central role in the articulation and meaning of the signs.

This assumption may be related, for example, to the definition of a new partitioning strategy that would allow more emphasis on the subtle traces of hands and fingers to the detriment of the other parts of the body. Also, the definition of specific weights for these parts would enable the model to learn more about its dynamics; today, the model does not distinguish the type of learned joint. Finally, including the depth information of coordinates may provide to the model more details about the trajectory of these parts, enabling the observation through the three-dimensional plane.